\documentclass[conference]{IEEEtran}
\IEEEoverridecommandlockouts
\usepackage{cite}
\usepackage{amsmath,amssymb,amsfonts}
\usepackage{algorithmic}
\usepackage{graphicx}
\usepackage{textcomp}
\usepackage{xcolor}
\usepackage{multirow}
\usepackage{booktabs}
\usepackage{lscape}
\usepackage{hyperref}

\def\BibTeX{{\rm B\kern-.05em{\sc i\kern-.025em b}\kern-.08em
    T\kern-.1667em\lower.7ex\hbox{E}\kern-.125emX}}
\begin{document}

\title{XQSV: A Structurally Variable Network to Imitate Human Play in Xiangqi}

\author{\IEEEauthorblockN{Chenliang Zhou}
\IEEEauthorblockA{\textit{Department of Computer Science and Technology} \\
\textit{University of Cambridge}\\
Cambridge, U.K. \\
chenliang.zhou@cl.cam.ac.uk}
}

\IEEEoverridecommandlockouts
\IEEEpubid{\makebox[\columnwidth]{ 979-8-3503-5067-8/24/\$31.00~\copyright2024 IEEE \hfill} 
\hspace{\columnsep}\makebox[\columnwidth]{ }}

\maketitle
\IEEEpubidadjcol
\begin{abstract}
In this paper, we introduce an innovative deep learning architecture, termed \emph{Xiangqi Structurally Variable (XQSV)}, designed to emulate the behavioral patterns of human players in Xiangqi, or Chinese Chess. The unique attribute of XQSV is its capacity to alter its structural configuration dynamically, optimizing performance for the task based on the particular subset of data on which it is trained. We have incorporated several design improvements to significantly enhance the network's predictive accuracy, including a local illegal move filter, an Elo range partitioning, a sequential one-dimensional input, and a simulation of imperfect memory capacity. Empirical evaluations reveal that XQSV attains a predictive accuracy of approximately 40\%, with its performance peaking within the trained Elo range. This indicates the model's success in mimicking the play behavior of individuals within that specific range. A three-terminal Turing Test was employed to demonstrate that the XQSV model imitates human behavior more accurately than conventional Xiangqi engines, rendering it indistinguishable from actual human opponents. Given the inherent nondeterminism in human gameplay, we propose two supplementary relaxed evaluation metrics. To our knowledge, XQSV represents the first model to mimic Xiangqi players.
\end{abstract}


\section{Introduction and Background}
\subsection{Board Game Engines}

The synergy between AI and board games has been particularly fruitful, leading to the development of powerful game engines that surpass human experts \cite{wiki2020topchess}. More recently, deep reinforcement learning has emerged as a powerful method including DeepMind's AlphaGo \cite{chaslot2008monte} and AlphaZero \cite{silver2016mastering, silver2017mastering, silver2018general} demonstrating superhuman performance in chess and Shogi, learning solely from game rules without supervised learning. Other notable engines include the chess engine TDLEAF \cite{baxter2000learning}, based on temporal difference learning, the Go engine based on deep convolutional neural networks \cite{tian2015better}, and the poker engine DeepStack, based on recursive reasoning \cite{moravvcik2017deepstack}.

\subsection{Imitate Human in Board Games}

Many studies focus on developing board game engines that optimize winning probability, but less attention is given to engines that imitate human behavior. Robust game performance does not equate to human-like play. For example, the Monte-Carlo Tree Search algorithm is inadequate for simulating human-like gameplay \cite{perez20152014}. Research shows that both Stockfish and Leela (an open-source AlphaZero implementation) struggle to predict human moves accurately, even when calibrated to match human performance \cite{mcilroy2020aligning}. This discrepancy arises from fundamental differences between machines and humans, such as machines' vast memory and processing capabilities. Another study found that AlphaZero prefers piece activity over material, favoring riskier, more aggressive actions \cite{kasparov2018chess}.

Creating an engine that performs in a human-like manner has multiple applications. It could serve as an independent game engine, offering a more enjoyable user experience, or assist human players with strategic advice. Playing alongside a human-oriented engine helps human players learn and collaborate more effectively. Additionally, such an engine can provide a robust evaluation function integrated into other game engines that prioritize optimal moves (e.g., \cite{greer2015more, xiao2016factorization}).

The common approach to emulate human-like behavior involves predicting human actions based on game history (e.g., \cite{greer2015more} for chess; \cite{xiao2016factorization} for Go; \cite{khalifa2016modifying} for video games). Maia \cite{mcilroy2020aligning}, based on AlphaZero \cite{silver2018general}, models human actions in chess in a detailed way. It is trained on human chess games to predict moves of players at specific skill levels, achieving 50\% prediction accuracy, surpassing Stockfish's 37\% and Leela's 42\%.

\subsection{Xiangqi Engines}

Xiangqi (Chinese chess) is a popular board game in China with a rich history dating back to the Warring States period (c. 475 B.C.). It simulates a battle on the chessboard, where players act as commanders to checkmate the opponent's king. Xiangqi offers a vast strategy space and a large game tree complexity ($10^{150}$, \cite{yen2004computer}), exceeding that of Chess ($10^{123}$, \cite{shannon1950xxii}). This complexity makes it a valuable platform for AI research, modeling reasoning and planning processes.

Most Xiangqi engines use reinforcement learning with self-play and game tree search along with other techniques \cite{he2019application, wang2009applying, fan2010research, du2014self, wang2016deep}. Despite extensive research, few studies focus on imitating human behavior in Xiangqi. Our paper is the first to address this gap, aiming to establish a benchmark and encourage further research in this field.

\subsection{Contributions}
The major contributions of our work are 1. \emph{Xiangqi Structurally Variable (XQSV)}, the pioneering neural network model adept at imitating the behaviors of human players in Xiangqi achieving an accuracy of 40\% and 2. an innovative scheme of variable network structure suitable for datasets that can be intuitively partitioned according to specific criteria.

\section{XQSV Design Approach}
\label{sec:XQSV Design Approach}
In order to imitate human behavior in Xiangqi, we formulate this problem as a classification problem over all Xiangqi moves.
In this section we highlight some innovative and essential design choices. 

\subsection{Data Preprocessing}
\label{sec:Data Preprocessing}
The raw Xiangqi game data is retrieved from \url{PlayOK.com} in standard algebraic notation (SAN) format. The data preprocessing workflow is as follows: partition Elo range, extract moves, group, map moves to numbers, break the moves into records, and finally, randomize the records. This section elaborates on the three critical preprocessing steps: partition Elo range, sequential 1D input, and imperfect memory capacity.


\subsubsection{Elo Range Partitioning}
\label{sec:Elo Range Split}
The Elo scoring system \cite{elo1967proposed} serves as a sophisticated metric for quantifying the relative skill levels of players in strategy-based games, including chess and Xiangqi. The variance observed within our dataset is, in part, attributable to the diverse skill levels of the players, ranging from novices to experts. Accordingly, partitioning the data based on Elo scores serves to mitigate this variability, consequently enhancing the model's ability to accurately learn human behavior.

After acquiring the raw dataset, we partitioned it into nine distinct bins based on the Elo scores of the participants, ranging from $(1000,1100]$ to $(1900,2000]$. This segmentation strategy stems from the inherent challenge posed by our predictive task, which is characterized by the substantial variability in players' abilities and performances juxtaposed against the deterministic predictions generated by our model. Ideally, the model's predictions would align with the moves most commonly executed by players. Nonetheless, if the dataset amalgamates players from a broad spectrum of skill levels, only a minority may exhibit the move deemed optimal by the model, thereby diminishing prediction accuracy.


\subsubsection{Sequential 1D Input}

Numerous existing game engines utilize data input in the form of two-dimensional chessboard configurations, a method potentially favored for its efficiency in representing game history (e.g., \cite{silver2016mastering, mcilroy2020aligning}). Contrarily, our XQSV anticipates a linear input in the form a sequential series of moves. This design decision is more closely aligned with the cognitive processing patterns observed in humans. Through the conduct of interviews with 30 Xiangqi players, we discovered that a predominant majority (28 out of 30) rely on the recollection of previous moves rather than the visual configurations of the chessboard to deliberate over the future moves. Given that XQSV's objective is to predict human movement, we hypothesize that it is beneficial if it operates more similarly to a human by considering the game history as a sequence of moves rather than chessboard configurations.

In the ablation study delineated in Section \ref{sec:ablation-study}, empirical evidence supports this hypothesis; the adoption of a one-dimensional move sequence as input, in lieu of two-dimensional chessboard configurations, markedly enhances the model’s predictive accuracy. Such findings suggest that for the purpose of predicting human movement within the game, a model that operates in a manner congruent with human cognitive patterns yields superior performance.

\subsubsection{Imperfect Memory Capacity}
Contrary to most existing game engines which operate under the assumption of a perfect memory that provides complete access to the entire game history, our XQSV model is restricted to a limited memory capacity, denoted as $m$. This means it can only consider at most $m$ moves from the past. This design decision has been adopted not only for computational efficiency, but more crucially, to better simulate the human cognitive process. It reflects the reality that during a game of Xiangqi, human players are typically unable to recall the complete history of the game owing to the inherent limitations of human brain capacity.

Formally, suppose a game history $H$ consists of a sequence of $l$ moves $H = (M_1, M_2, \ldots, M_l)$, then we will produce $l$ training samples $(x,y)$ from $G$ by going through each step while maintaining the number of history steps under $m$:
\begin{align}
    \label{eq:truncate}
    \left(x=M_{\max(1,i-m)...i-1}, y=M_i \right), \; i=1,\ldots,l ,
\end{align}
where $M_{j...k}$ is the subsequence $(M_j, M_{j+1}, \ldots, M_k).$





One might argue that artificially limiting the machine's memory is unnecessary; instead, we could supply the complete game history and allow the model to autonomously discern which move to discard. For example, long short-term memory (LSTM) \cite{hochreiter1997long} or gated recurrent unit (GRU) \cite{cho2014learning} have forget gates integrated into their network for this purpose. However, we note that the effective functioning of these forget gates relies upon a successful learning of the underlying data structure, which may not always be the case, particularly considering the complexity of our prediction task. 

\subsection{Structurally Variable RNN}
\label{structure}
XQSV employs sequential modeling and is constructed utilizing a recurrent neural network (RNN) followed by fully connected layers (FC). The network calculates the probability distribution for each move and selects the move with the highest probability. Numerous chess engines are based on convolutional neural network (CNN) or residual neural network (ResNet) (e.g., \cite{silver2016mastering, mcilroy2020aligning}. However, given that our input consists of a sequence of 1D data, we chose to utilize RNN as the primary architecture. Our preliminary experiments indicated that a coarsely tuned LSTM \cite{hochreiter1997long} could already achieve a prediction accuracy comparable to that of CNN or ResNet. Furthermore, RNN assumes dependencies among elements within the input sequence, a condition which is applicable to sequential Xiangqi moves and makes the model better resemble the cognitive process of human brain.

A notable innovation of XQSV resides in its variable network structure. As previously mentioned in Section \ref{sec:Elo Range Split}, we partitioned the dataset according to different Elo ranges to minimize intra-group variation. For each Elo range, a distinctive architecture is required as players at various skill levels are likely to exhibit divergent thought processes and behaviors. For instance, it is suggested that advanced players possess a superior capability to reconstruct a chessboard situation, attributed to their capacity to encode the chessboard into larger perceptual chunks using more abstract relations \cite{chase1973perception}. Consequently, to imitate these players, a model with a more complex structure may be required, a hypothesis reaffirmed by our experiments in \ref{sec:experiments}. Thus, we introduced variability into the network structure, allowing it to adapt autonomously to players at different skill levels. In this setting, the model must not only learn to optimize network weights but also, crucially, determine the most efficacious network architecture at a higher level.

Several control elements, designated as structure variables (SVs), were established to manipulate the variable structure of the network. Ten SVs are summarized in Table~\ref{tab:xqsv-candidate-values}. Among these, the memory capacity $m$ determines the number of past moves the model can consider, with different Elo ranges potentially requiring different optimal $m$ values. This structurally variable framework can be generalized to accommodate other tasks, particularly when there is an intrinsic partitioning of the training dataset that can influence the output.

\begin{table*}[ht]
\small
\centering
\caption{Structural variables and their candidate values}
\label{tab:xqsv-candidate-values}
\begin{tabular}{c|c|c}
\toprule
Structural variable & Abbr. & Candidate values (\textbf{default in bold})                      \\ \midrule
Memory Capacity & $m$ & \textbf{5}, 10, 15, 20 \\ 
Concrete RNN layer & RNN          & \textbf{LSTM}, GRU, Backward LSTM, Barkward GRU \\ 
RNN dropout probability & RDP       & 0, \textbf{0.05}, 0.1,0.2                           \\ 
RNN hidden dimension & RHD & \textbf{512}, 1024, 2048                        \\ 
RNN activation function & RA     & \textbf{ReLU}, Softmax, Linear, Tanh                \\ 
Batch normalization layer & BN  & \textbf{Yes}, No                                \\ 
Dropout layer probability & DP & 0, \textbf{0.05}, 0.1,0.2                               \\ 
Number of fully connected layers & \#FC  & 0, 1, \textbf{2}, 3, 5                                \\ 
FC regularization coefficient & FCR          & 0, \textbf{0.001}, 0.002, 0.005                 \\ 
FC activation function & FCA         & ReLU, \textbf{Softmax}, Linear, Tanh                \\ \bottomrule
\end{tabular}
\end{table*}


\subsection{Locally Illegal Move Filter}
\label{sec:filter}
One challenge in the move prediction task resides in the large search space.
After carefully considering the rules of Xiangqi for each game piece, we successfully condensed the label space to a minimal set of 755 moves. These moves are referred to as \emph{globally legal moves}.

To further facilitate the prediction task, we observed that these globally legal moves are not always legal given a specific board configuration. Thus, we introduce the concept of a \emph{locally legal move}, which imposes stricter conditions than a global legal move: a move must not only abide by piece movement rules but also be legally executable given a specific board arrangement. For example, if a piece occupies a certain position, all moves moving the piece from a different position are deemed locally illicit. 
Upon examining the output of XQSV in pilot experiments, we observed that certain moves with high probabilities in the output were, in fact, not locally legal. Such prediction errors can be circumvented through the implementation of a \textit{locally illegal move filter}. We report the improvement brought by this filter in Section \ref{sec:ablation-study}.

\section{Experiment and Evaluation}
\label{sec:Experiment and Evaluation}

\subsection{Training XQSV on six Elo ranges}
\label{sec:experiments}
\begin{figure}[tb]
    \centering
    \includegraphics[width=0.4\textwidth]{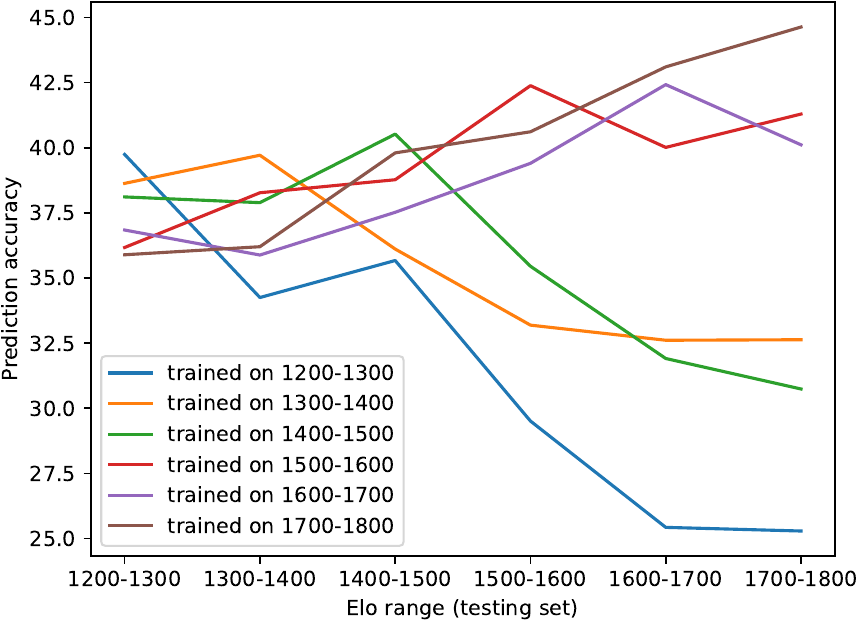}
    \caption{XQSV's prediction accuracy on different Elo ranges}
    \label{fig:compare}
\end{figure}

We illustrate the effectiveness, specifically the adaptability, of XQSV through its training on six Elo ranges: $(1200,1300], (1300,1400], \ldots, (1700,1800]$. Given the computational constraints and that there are ten structural variables, an exhaustive search for the optimal network structure was not feasible. Instead, we implemented a five-phase search, with each phase refining the outcome of the preceding phase. During each phase, we sought the optimal values for two structural variables, in the order presented in Table~\ref{tab:xqsv-candidate-values}. For the remaining structural variables, we either utilized their optimal values found in previous phases, or, if they have not been searched, the default values as shown in the table. The best network structures identified for each Elo range are reported in Table~\ref{tab:exp-result}. We also evaluated each model on different Elo ranges in addition to the one it was trained on and visualized the results in Fig.~\ref{fig:compare}.


\begin{table*}[ht]
\small
\centering
\caption{Best SV configuration for each Elo range}
\label{tab:exp-result}
\begin{tabular}{c|c|c|c|c|c|c|c|c|c|c|c}
\toprule
Elo Range & Acc. (\%) & $m$ & RNN & RDP & RHD & RA & BN & DP & \#FC & FCR & FCA  \\ \midrule
1200-1300 & 39.74 & 10 & GRU & 0.1 & 1024 & \multirow{6}{*}{ReLU} & \multirow{6}{*}{Yes} & \multirow{6}{*}{0.05} & 1 & 0.002 & \multirow{6}{*}{Softmax} \\ 
1300-1400 & 39.71 & 10 & GRU & 0.1 & 512 & & & & 1 & 0.002 &  \\  
1400-1500 & 40.52 & 10 & LSTM & 0.1 & 512 & & & & 1 & 0.001 &  \\ 
1500-1600 & 42.38 & 20 & GRU & 0.05 & 512 & & & & 1 & 0.001 &  \\ 
1600-1700 & 42.42 & 20 & LSTM & 0.05 & 1024 & & & & 2 & 0.001 &  \\ 
1700-1800 & 44.63 & 20 & LSTM & 0.05 & 1024 & & & & 2 & 0.001 &  \\ \bottomrule
\end{tabular}

\end{table*}

Table~\ref{tab:exp-result} clearly shows that as the skill level of the player increases, a more complex network is required to imitate their gameplay. This is reflected through the larger memory capacity, a more advanced RNN layer (LSTM being more complex than GRU), reduced regularization, increased RNN hidden dimension, and a higher number of fully connected layers. This aligns with findings from cognitive psychology research (e.g., \cite{chase1973perception}) which suggests that more advanced Xiangqi players exhibit different cognitive strategies and tend to have superior recall capabilities (hence the larger $m$) and the ability to abstract the chessboard into cognitive chunks (necessitating a more complex network).

Fig.~\ref{fig:compare} demonstrates that when tested on game records across various Elo ranges, all XQSVs achieve peak accuracy in the Elo range on which they were trained. As the Elo range deviates from this, the prediction accuracy typically declines. This clearly indicates that XQSV has effectively learned the specific playing patterns and characteristics of the Elo range it was trained on. Interestingly, we note that the decline in accuracy is less pronounced for XQSVs trained on higher Elo ranges. This could be interpreted as a reflection of advanced players' ability to emulate novice strategies, but not vice versa. In other words, certain moves may occur across all skill levels, while more sophisticated, nuanced moves are exclusive to advanced game records.



\subsection{Ablation Study}
\label{sec:ablation-study}
In this section, we perform an ablation study to examine the effectiveness of the four key design decisions in XQSV architecture, which were proposed to improve prediction accuracy in Sections \ref{sec:Data Preprocessing} and \ref{sec:filter}:
\begin{itemize}
    \item \textbf{Elo range partitioning}. To ablate this component, we rerun the experiment trained on the whole dataset of game records without Elo range partitioning.
    
    \item \textbf{Sequential 1D input}. To ablate this component, we encode the game history as a sequence of 2D chessboard situations. Instead of RNN, we use a ResNet built upon CNN to process the 2D input followed by fully connected layers to output the predicted move. This is the approach taken in most of the chess engines (e.g., \cite{silver2016mastering, silver2017adrian, silver2018general, mcilroy2020aligning}). For a fair comparison, we also turn this network into a structurally variable one. Most SVs in Table~\ref{tab:xqsv-candidate-values} not related to RNN are reusable, and in addition, we introduce two SVs for the ResNet, namely the number of convolutional blocks and the number of channels in convolutional layers.
    
    \item \textbf{Imperfect memory capacity}. To ablate this component, we  rerun the experiments on the game records without truncation on the moves in the far past.

    \item \textbf{Locally illegal move filter}. To ablate this component, we rerun the experiment but removing the locally illegal move filter module.
\end{itemize}

When rerunning the experiments, we perform the same multi-phase adapting and training procedure described in Section \ref{sec:experiments}; we train for the same number of epochs until convergence at the end of training. The result is reported in Table~\ref{tab:ablation-study}.



\begin{table*}[ht]
\small
\centering
\caption{Prediction accuracy (\%) for ablation study }
\label{tab:ablation-study}
\begin{tabular}{c|c|c|c|c|c}
\toprule
Elo Range & Ours &  No Elo partitioning & 2D input (ResNet) & Perfect memory & No illegal move filter \\ \midrule
1200-1300 & \textbf{39.74} & \multirow{6}{*}{18.42} & 33.18 & 24.52 & 24.43        \\ 
1300-1400 & \textbf{39.71} & & 33.83 & 27.56 & 25.64 \\
1400-1500 & \textbf{40.52} & & 34.64 & 32.65 & 23.94 \\
1500-1600 & \textbf{42.38} & & 35.68 & 34.27 & 24.50 \\
1600-1700 & \textbf{42.42} & & 37.11 & 36.47 & 25.28 \\
1700-1800 & \textbf{44.63} & & 35.58 & 33.23 & 23.58 \\ \bottomrule
\end{tabular}
\end{table*}

We can observe from the table that all of the four design decisions above are essential to our model and removing any of them caused significant decrease in the prediction accuracy. Additionally, it is worth noting that the computational time increased substantially when employing an unlimited memory capacity.

\subsection{Turing Test}
We also performed a three-terminal Turing Test \cite{turing1950computing} to determine whether human Xiangqi players can distinguish our XQSV from a true human opponent. We invited 30 Xiangqi players with Elo scores between 1200-1500. Each participant was asked to remotely play nine Xiangqi games with three different opponents in random order consisting of the following:
\begin{itemize}
    \item three games with a human opponent randomly selected from the remaining 29 participants;
    \item three games with our XQSV trained at Elo range 1300-1400; and
    \item as a control group, three games with a mobile App called \textit{Chinese Chess} downloaded from Apple's App Store.
\end{itemize}
The participants were aware of the above rules, and knew that which set of three games was played with the same opponent. However, they were not informed about the identities of the human and machine opponents (playing remotely allowed to hide this information from them). After each game, the participants were asked to identify which of the three opponents they believed to be human (thereby indicating the remaining two as machines)

In this setup, each of the three opponents received a total of $30 \times 3 = 90$ guesses of being human. Among these, both the human player and our XQSV were identified as human 35 times (38.89\%), while the \textit{Chinese Chess} App 20 times (22.22\%). Note that a higher percentage score means that the opponent behaves more like a human.

From the results, we see that $77.78\%$ of guesses correctly identified the \emph{Chinese Chess} App as a computer program, indicating that it, and potentially other popular Xiangqi game engines, did not perform well in imitating human moves and could provide a different gaming experience from that of playing with genuine human opponents; In contrast, our XQSV model demonstrated superior proficiency in imitating human moves, outperforming the \emph{Chinese Chess} App by a moderate margin. Over the course of these 90 Xiangqi games, our XQSV model was indistinguishable from the actual human opponent, indicating that XQSV can successfully imitate human moves in Xiangqi.

\section{Deterministic Model vs. Non-Deterministic Human Play}
\label{sec:Deterministic Model vs. Non-Deterministic Human Play}

It is important to note that a significant challenge in this prediction task stems from the non-deterministic nature of human play, in contrast to the deterministic nature of our model: In the face of identical chessboard situations, multiple moves could be selected by various players, or even by the same player at different times, but our model will always output the same move. While we have attempted to address this issue through data preprocessing by partitioning the data according to Elo scores, this approach cannot eliminate the non-determinism inherent to the game

To account for this, recall that before predicting a single move, our model already computes a probability distribution over all possible moves. To incorporate the non-determinism of human-play, we can randomize the prediction by stochastically outputting each move according to the computed probability. 

Given the inherent non-determinism of human play, there are inevitable ``unfair misses'' during the prediction. For instance, in a given chessboard situation, there may be five equally good moves. Given our model's deterministic nature, it can only predict one of those moves, capping the maximum prediction accuracy at 20\%. This accuracy diminishes rapidly as the number of equally good moves increases, a common scenario in complex game situations. Consequently, it's pertinent to introduce alternative evaluation metrics to more accurately capture our model's performance.

One such metric can be the top $k$ accuracy. In this approach, instead of requiring the predicted move to be the one with the highest probability in the output distribution, it just needs to be within the top $k$ moves with the highest probabilities. Note that a larger $k$ corresponds to a less strict criteria. Under this metric, a predicted move $M$ is considered correct as long as it's frequently chosen by human players, not necessarily being the most common choice. This approach accommodates the game's intrinsic non-determinism by acknowledging that there can be multiple moves with high probabilities. Considering that we have 755 moves to predict, we believe this is a reasonable and practical adjustment to our evaluation metric.

The top $k$ accuracy  assumes that in every situation there are $k$ good moves, but sometimes the number of good moves is unpredictable and may vary significantly. To better reflect the inherent variability, we also propose the ``top $p$ probability accuracy'', where a prediction is considered correct if it is among the minimal set $S$ of moves with the highest probabilities such that the sum of their probabilities exceeds $p$. In other words, we form a set of good moves by ordering all moves from highest to lowest probability, then sequentially add moves to this set until the cumulative probability exceeds the threshold $p$. This approach allows for a more accurate evaluation of model performance. For a given threshold $p$, the size of the set $S$ is dynamically adjusted: If there is a single correct move $M$, its probability will likely be high, and the set $S$ will consist solely of this move. Conversely, if there are multiple equally good moves, each will likely have a smaller individual probability, and $S$ will contain all of these moves. This flexible approach more accurately captures the intricacies and complexities of Xiangqi gameplay.

We reevaluate the model trained on Elo range 1200-1300 on the testing dataset with Elo range 1200-1300. In Table~\ref{tab:top-k-acc}, we report the top $k$ accuracy and top $p$ probability accuracy for different $k$ and $p$. Notice that the top 1 accuracy and the top 0 probability accuracy are just the ordinary (strictest) accuracy we have used in previous sections. 

\begin{table}[ht]
\small
\centering
\caption{Evaluating XQSV on Elo range 1200-1300 with relaxed metrics}
\label{tab:top-k-acc}
\begin{tabular}{c|c|c|c}
\toprule
$k$ & Top-$k$ acc. (\%) & $p$ & Top-$p$ prob. acc. (\%) \\ \midrule
1  & 39.74 & 0   & 39.74 \\ 
2  & 47.82 & 0.1 & 41.51 \\ 
3  & 53.68 & 0.2 & 43.71 \\ 
4  & 58.37 & 0.3 & 48.26 \\ 
5  & 62.30 & 0.4 & 55.39 \\ 
6  & 65.61 & 0.5 & 63.18 \\ 
7  & 68.59 & 0.6 & 74.72 \\ 
8  & 71.05 & 0.7 & 85.62 \\ 
9  & 73.33 & 0.8 & 93.12 \\ 
10 & 76.38 & 0.9 & 97.38 \\
755 & 100.00 & 1.0 & 100.00 \\ \midrule
\end{tabular}
\end{table}

As expected, as $k$ and $p$ increase, the prediction accuracy generally increases as well. This is because as we expand the pool of acceptable predictions (by including more top moves or those that have a higher cumulative probability). If we make a reasonable assumption that given a chessboard situation, there are about five moves that could be considered good by human, then the accuracy of XQSV adapted to Elo range 1200-1300 is 62.30\%.

\section{Discussion and Conclusion}


One limitation of our work lies in the discord between non-deterministic human behavior and deterministic model prediction. We attempted to counterbalance this by partitioning the data according to Elo ranges and applying a relaxed accuracy metric. However, the model could potentially benefit from other effective methodologies. For instance, predicting a player's playing style or personal traits - either via a separate or integrated network, or other methods - could further facilitate the prediction of human moves.


To conclude, in this paper we devised an innovative model termed XQSV (Xiangqi Structurally Variable), which employs a recurrent neural network (RNN) and dynamically alters its network structure to optimally represent Xiangqi players across different proficiency levels. We further introduced four key design approaches that substantially enhanced the predictive accuracy of our model: a local illegal move filter, an Elo range partitioning, a sequential 1D input, and an imperfect memory capacity. Evaluated on Xiangqi game data, XQSV demonstrated an approximate accuracy of 40\%, which, under justifiable relaxation of evaluation metrics, increased to around 60\%. Through extensive experimentation, we showed the ability of XQSV to effectively imitate human players. To the best of our knowledge, XQSV is the first model specialized in replicating human behaviour in Xiangqi, and we hope to establish a valuable benchmark for future endeavours in this research field.

\bibliographystyle{IEEEtran}
\bibliography{references}
\end{document}